\documentclass[sigconf, nonacm, protrusion = true, final]{acmart}

\usepackage{multirow}
\usepackage{comment}
\usepackage{subfigure}
\usepackage{enumitem}
\usepackage{hyperref}
\usepackage{balance}
\usepackage{csquotes}
\usepackage{tikz}
\usepackage{bbding}
\allowdisplaybreaks[4]
\newcommand*\circled[1]{\tikz[baseline=(char.base)]{
            \node[shape=circle,draw,inner sep=2pt,scale=0.6] (char) {#1};}}

\newcommand\vldbdoi{XX.XX/XXX.XX}
\newcommand\vldbpages{XXX-XXX}
\newcommand\vldbvolume{14}
\newcommand\vldbissue{1}
\newcommand\vldbyear{2020}
\newcommand\vldbauthors{\authors}
\newcommand\vldbtitle{\shorttitle} 
\newcommand\vldbavailabilityurl{}
\newcommand\vldbpagestyle{plain} 

\begin{document}
\title{SQLfuse: Enhancing Text-to-SQL Performance through Comprehensive LLM Synergy}

%%
%% The "author" command and its associated commands are used to define the authors and their affiliations.
\author{Tingkai Zhang}
\authornote{Equal contribution.}
\affiliation{%
    \institution{Ant Group}
    \country{China}
  }
\email{tingkai.ztk@antgroup.com}

\author{Chaoyu Chen}
\authornotemark[1]
\affiliation{%
    \institution{Ant Group}
    \country{China}
  }
\email{chris.ccy@antgroup.com}

\author{Cong Liao}
\authornotemark[1]
\affiliation{%
    \institution{Ant Group}
    \country{China}
  }
\email{liaocong.lc@antgroup.com}

\author{Jun Wang}
\authornotemark[1]
\affiliation{%
    \institution{Ant Group}
    \country{China}
  }
\email{wangjun412042@antgroup.com}

\author{Xudong Zhao}
\affiliation{%
    \institution{Ant Group}
    \country{China}
  }
\email{gordon.zxd@antgroup.com}

\author{Hang Yu}
\authornote{Corresponding author.}
\affiliation{%
    \institution{Ant Group}
    \country{China}
  }
\email{hyu.hugo@antgroup.com}

\author{Jianchao Wang}
\authornotemark[2]
\affiliation{%
    \institution{Ant Group}
    \country{China}
  }
\email{luli.wjc@antgroup.com}

\author{Jianguo Li}
%\authornotemark[2]
\authornotemark[2]
\affiliation{%
    \institution{Ant Group}
    \country{China}
  }
\email{lijg.zero@antgroup.com}

\author{Wenhui Shi}
\authornotemark[2]
\affiliation{%
    \institution{OceanBase, Ant Group}
    \country{China}
  }
\email{yushun.swh@oceanbase.com}

% \renewcommand{\shortauthors}{Chen et al.}
%\renewcommand{\shortauthors}{Zhang et al.}

\begin{abstract}
% Text-to-SQL 是自然语言处理领域的一个重要研究方向，它旨在将用户的自然语言查询转换成可执行的SQL语句。近期研究表明，在这一任务上，基于GPT的模型表现出色。GPT以其强大的语言生成能力，在理解自然语言和生成准确的SQL语句方面取得了显著的进展。特别是在一些公认的基准数据集上，GPT模型不仅显著超越了传统的基于规则和基于模板的方法，而且在某些情况下，它的表现甚至接近专业数据库管理员手写的查询效果。

% 但在Text-to-SQL任务中，关于开源大型语言模型的研究工作相对较少，并且这些模型在该任务上的性能水平尚未被广泛验证。大部分现有研究依赖于私有或者定制化模型，这些模型可能进行了特别的预训练和微调以适应转换自然语言查询到SQL的特定需求。因此，关于如何有效利用开源LLMs进行Text-to-SQL转换，以及这些模型在不同类型和规模的数据库上的适用性和性能表现，仍有大量的探索空间。

% 为了充分发掘开源大规模语言模型（LLMs）在处理Text-to-SQL任务中的潜力，本文深入探讨了它们在多种不同的Text-to-SQL场景下的性能。我们针对各种数据库上下文和查询难度，对开源LLMs进行了广泛的基准测试，旨在揭示模型在处理各种自然语言问询到SQL语句转换任务时的能力。此外，本研究进一步通过有监督的微调方法对LLMs进行优化，细致调整模型参数，在 Spider \cite{yu2018spider} 的测试集上，执行准确度为 85.6，成为榜单中最佳开源模型，略低于基于大量 Few-shot 的 GPT-4 方案，高于其他 GPT-3.5/4 方案

Text-to-SQL conversion is a critical innovation, simplifying the transition from complex SQL to intuitive natural language queries, especially significant given SQL's prevalence in the job market across various roles. The rise of Large Language Models (LLMs) like GPT-3.5 and GPT-4 has greatly advanced this field, offering improved natural language understanding and the ability to generate nuanced SQL statements. However, the potential of open-source LLMs in Text-to-SQL applications remains underexplored, with many frameworks failing to leverage their full capabilities, particularly in handling complex database queries and incorporating feedback for iterative refinement. Addressing these limitations, this paper introduces SQLfuse, a robust system integrating open-source LLMs with a suite of tools to enhance Text-to-SQL translation's accuracy and usability. SQLfuse features four modules: schema mining, schema linking, SQL generation, and a SQL critic module, to not only generate but also continuously enhance SQL query quality. Demonstrated by its leading performance on the Spider Leaderboard and deployment by Ant Group, SQLfuse showcases the practical merits of open-source LLMs in diverse business contexts.

\end{abstract}

\maketitle

\pagestyle{\vldbpagestyle}
\begingroup\small\noindent\raggedright\textbf{PVLDB Reference Format:}\\
\vldbauthors. \vldbtitle. PVLDB, \vldbvolume(\vldbissue): \vldbpages, \vldbyear.\\
\href{https://doi.org/\vldbdoi}{doi:\vldbdoi}
\endgroup
\begingroup
\renewcommand\thefootnote{}\footnote{\noindent
This work is licensed under the Creative Commons BY-NC-ND 4.0 International License. Visit \url{https://creativecommons.org/licenses/by-nc-nd/4.0/} to view a copy of this license. For any use beyond those covered by this license, obtain permission by emailing \href{mailto:info@vldb.org}{info@vldb.org}. Copyright is held by the owner/author(s). Publication rights licensed to the VLDB Endowment. \\
\raggedright Proceedings of the VLDB Endowment, Vol. \vldbvolume, No. \vldbissue\ %
ISSN 2150-8097. \\
\href{https://doi.org/\vldbdoi}{doi:\vldbdoi} \\
}\addtocounter{footnote}{-1}\endgroup

\ifdefempty{\vldbavailabilityurl}{}{
\vspace{.3cm}
\begingroup\small\noindent\raggedright\textbf{PVLDB Artifact Availability:}\\
The source code, data, and/or other artifacts have been made available at \url{\vldbavailabilityurl}.
\endgroup
}

\section{Introduction}

% 在 2023 年 IEEE Spectrum 发布的《第十届年度顶级编程语言榜单》中，SQL 在 Jobs 榜单继续蝉联第一。SQL对 BI、DEV、DBA 等专业人士来说是必备的技能，甚至产品、运营、合规、商业决策者等都会涉猎一二，几乎任何需要数据的场景都可能用到 SQL。

% 但在传统场景中，用户通常需要具有一定的数据库和 SQL 语言知识，并需要理解数据存储，才能准确地编写 SQL 查询语句。正是在这种背景下，Text-to-SQL 的技术显示出其巨大的价值和潜力。

% Text-to-SQL 可以将原先使用复杂的 SQL 语言查询数据的方式，转化为更自然、更具可读性的自然语言进行，让用户轻松地查询数据，分析查询结果。它填补了非专业用户与数据库系统之间的空白，大大提高了数据处理的效率，并有助于拓展应用范围，如数据库问题解答、SQL 调优等。然而，早期的研究将 Text-to-SQL 视为一个序列到序列的任务，重点在于使用编码器-解码器架构训练机器学习模型。此外，为了缩小  

% Text-to-SQL 研究与实际应用之间的差距，业界发布了许多权威的大规模基准数据集，包括 WikiSQL，Spider，KaggleDBQA，BIRD 等。随着 GPT 的发展， large language models (LLMs) 逐渐成为 Text-to-SQL 任务新的 paradigm，LLMs 进一步增强了自然语言处理的能力，使得 Text-to-SQL 任务更加准确和用户友好。这些模型以其巨大的参数量和广泛的训练数据集获得了对自然语言更深刻的理解，使得它们能够更好地解析和理解用户的查询意图，从而生成更准确的 SQL 代码。在这样的发展背景下，Text-to-SQL 技术不仅能够处理结构化查询，而且开始能够处理更复杂的多步骤查询、嵌套查询和具有上下文依赖的查询。此外，LLMs 的优势在于其能够继续学习和适应新的数据集和查询类型，即使在部署后也可以通过用户的反馈不断优化和改进模型的性能。

% 为了进一步提升 Text-to-SQL 的实用性和可靠性，研究者们开始集中精力解决一些关键问题：

% 1. 语义理解：LLMs 需要通过对自然语言的深层语义理解，进而更好地映射复杂的查询到相应的 SQL 语句。

% 2. 上下文感知：模型需要理解查询的上下文，包括之前的交互历史，以生成连贯和相关的 SQL 语句。

% 3. 域适应性：LLMs 需要适应不同的领域和行业的数据，灵活处理各种专业术语和数据架构。

% 4. 交互式查询优化：通过与用户的交互，LLMs 能够更好地理解用户的具体需求，并提供更加优化的查询结果。

% 5. 错误检测与自我修正：模型需要能够检测潜在的错误，并在必要时提出改正方案或索取用户的进一步指导。

% 鉴于上述挑战，我们着手研究并充分发掘了开源大规模语言模型（LLMs）在处理这些复杂自然语言理解任务上的内在潜能。在我们的研究中，我们专注于探索如何简化Text-to-SQL转换流程，将其分解为若干子模块，以提高整体系统的灵活性。我们致力于通过实验不同的提示（prompts）和数据表示形式，来找到能够最有效激发开源大规模语言模型（LLMs）潜力的最优实践方法。具体来讲，我们尝试了多种prompt工程策略，包括自然语言指示和结构化提示，以及不同的数据预处理和编码方式，旨在减少模型对输入数据的误解并提升其对SQL语言结构的生成准确性。我们旨在揭示开源LLMs在Text-to-SQL转换方面的强大功能，并探讨如何通过训练策略和模型调整来进一步激发它们的潜力。这一研究工作不仅为理解和优化开源LLMs在专业应用场景下的表现提供了宝贵的见解，同时也为未来研究如何将这些开源模型应用于其他自然语言处理任务提供了可行的路径。

In the ``Top 10 Programming Languages'' annual report by IEEE Spectrum for 2023, SQL maintains its top rank on the ``Jobs list'', highlighting the enduring demand for SQL proficiency in the job market\cite{topprogramminglanguages2023}. SQL's ubiquity is evident across a myriad of professional roles, including Business Intelligence (BI) analysts, developers, database administrators (DBAs), and even extends to product management, operations, compliance, and business strategy, owing to its critical function in data-driven decision-making. Nevertheless, mastery of SQL requires a deep understanding of database structures and the language itself, constituting a barrier for those without technical expertise.

Text-to-SQL technology represents a pivotal breakthrough, simplifying the process of data querying from the intricate SQL to a more intuitive natural language format. This innovation facilitates user-friendly data interrogation and analysis, democratizing access to database systems and thereby enhancing data processing efficiency and broadening its applications. In response to this practical need, the field has seen the release of substantial benchmark datasets such as WikiSQL \cite{yin2020tabert}, Spider \cite{yu2018spider}, KaggleDBQA \cite{lee2021kaggledbqa}, and BIRD \cite{li2024can}. These resources serve to bridge the gap between academic research and the tangible needs of the industry, fostering developments that are firmly rooted in practical application.

\begin{table*}[t]
\centering
\caption{Moduels and tools used in different related works (\Checkmark for yes and \XSolidBrush for no).} 
% \vspace{-2ex}
\label{tab:related-works-methods}
\resizebox{\textwidth}{!}{
\begin{tabular}{|c|cccc|cc|ccc|cc|}
\hline
 \multirow{2}*{Solution}  & \multicolumn{4}{c|}{\bf Schema Mining} & \multicolumn{2}{c|}{\bf Schema Linking} & \multicolumn{3}{c|}{\bf SQLgen} & \multicolumn{2}{c|}{\bf Critic Module} \\
\cline{2-12}
 & Primary Key    & Foreign Key   & One-to-Many  & Enumerations   & Extraction   & Ranking & CoT & SQL Execution Checking   & Constant Value Fix  & Critic Model    & Knowledge Base \\
\hline
 DAIL-SQL~\cite{gao2023texttosql}       &\Checkmark&\Checkmark&\XSolidBrush&\XSolidBrush &\XSolidBrush&\XSolidBrush &\XSolidBrush&\XSolidBrush&\XSolidBrush &\XSolidBrush&\Checkmark   \\
 DIN-SQL~\cite{pourreza2024din}        &\Checkmark&\Checkmark&\XSolidBrush&\XSolidBrush &\Checkmark&\XSolidBrush &\Checkmark&\Checkmark&\XSolidBrush &\XSolidBrush&fixed quantity   \\
 C3~\cite{dong2023c3}           &\Checkmark&\Checkmark&\XSolidBrush&\XSolidBrush &\Checkmark&\Checkmark &\XSolidBrush&\XSolidBrush&\XSolidBrush &\XSolidBrush&\XSolidBrush   \\
 RESDSQL-3B~\cite{li2023resdsql}     &\Checkmark&\Checkmark&\XSolidBrush&\XSolidBrush &\Checkmark&\Checkmark &\XSolidBrush&\XSolidBrush&\XSolidBrush &\XSolidBrush&\XSolidBrush   \\
 SQL-PaLM~\cite{{sun2023sql}}       &\Checkmark&\Checkmark&\XSolidBrush&\XSolidBrush &\XSolidBrush&\XSolidBrush &\XSolidBrush&\XSolidBrush&\XSolidBrush &\XSolidBrush&\XSolidBrush   \\
 SQLfuse        &\Checkmark&\Checkmark&\Checkmark&\Checkmark &\Checkmark&\Checkmark &\Checkmark&\Checkmark&\Checkmark &\Checkmark&\Checkmark   \\
\hline
\end{tabular}}
% \vspace{-2ex}
\end{table*}

With the advent of GPT-3.5 and GPT-4, Large Language Models (LLMs) have emerged as a transformative force for Natural Language Processing (NLP) tasks, Text-to-SQL included. These models' expansive parameters and rich training data have culminated in a nuanced grasp of natural language, yielding more accurate SQL translations by parsing user intents effectively. Despite these advancements, current LLM-based Text-to-SQL frameworks are not fully optimized; they do not exhaust the possibilities offered by open-source LLMs, nor do they effectively incorporate external tools and knowledge that could refine Text-to-SQL performance. Our analysis, detailed in Table~\ref{tab:related-works-methods}, indicates that current systems are limited in their approach. For instance, they often overlook complex one-to-many relationships crucial for constructing aggregate queries, such as totals, averages, and counts. Furthermore, they generally do not capitalize on execution error feedback, which can provide valuable insights for correcting SQL inaccuracies. Additionally, they lack a critic module designed to evaluate and rank SQL outputs from LLMs, which could significantly improve results.

\begin{figure*}
    \centering
    \includegraphics[width=1\linewidth]{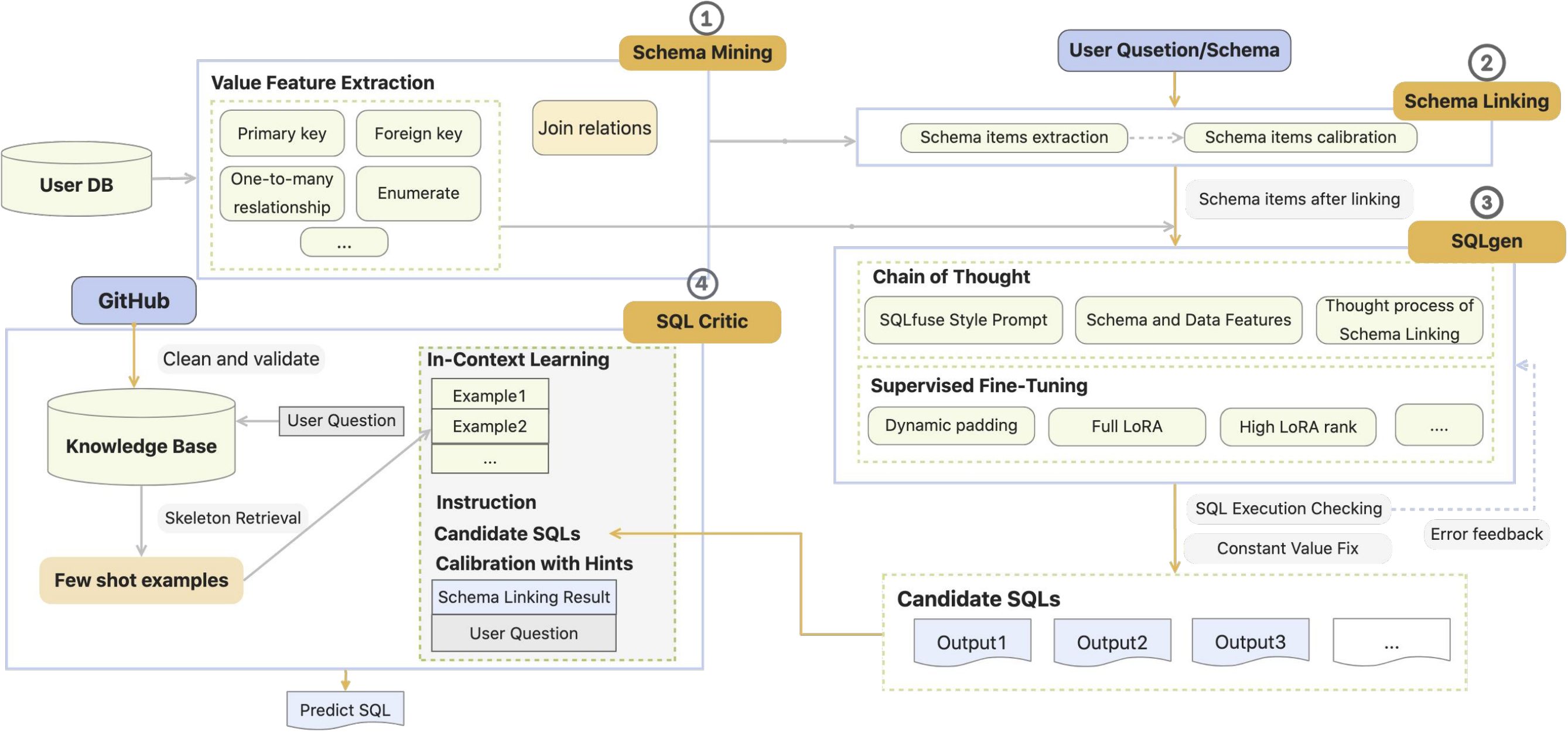}
    % \includegraphics[width=\textwidth]{image.png}
    %\vspace{-1ex}
    \caption{SQLfuse Architecture: \protect\circled{1} Schema Mining, \protect\circled{2} Schema Linking, \protect\circled{3} SQLgen, \protect\circled{4} SQL critic.}
    \label{fig:architecture-diagram}
    %\vspace{-1ex}
\end{figure*}

This paper introduces SQLfuse, a comprehensive system designed to harness the full spectrum of open-source LLMs, alongside a suite of tools and collective knowledge. As shown in Figure~\ref{fig:architecture-diagram}, SQLfuse is structured into four synergistic modules: schema mining, schema linking, SQL generation (SQLgen), and SQL critic. The schema mining module delves into the database to extract important keys and enumeration values, as well as their intricate relationships across tables. Schema linking then cohesively integrates these discoveries with user queries to form a logical sequence of thoughts, setting the stage for subsequent processes. Using this foundation, SQLgen employs fine-tuned LLMs to craft SQL queries that accurately reflect user intentions. To refine the queries after generation, a constant check module is employed and execution feedback is utilized for progressive improvements. Finally, the critic module employs an external database of quality queries to discern and select the most effective SQL output generated by SQLgen. The holistic integration of SQLfuse has not only placed it at the forefront of the Spider Leaderboard—with an impressive 85.6\% accuracy, ranking the first in the open-source LLM category and claiming the fourth position overall—but it has also proven its mettle in real-world applications. Ant Group has deployed SQLfuse across its platforms, utilizing the system across a range of business contexts, including its preeminent online data analytical processing and transaction processing platforms. In summary, our contributions include:
\begin{itemize}%[leftmargin=1.5em,itemsep= 2pt,topsep = 2pt,partopsep=2pt]
    \item  We introduce SQLfuse, an innovative and comprehensive Text-to-SQL system developed with the integration of open-source LLMs. SQLfuse is architecture around four synergistic modules: schema mining, schema linking, SQLgen, and SQL critic, each designed to harness and amplify the capabilities of LLMs for enhanced query translation accuracy.
    \item Through rigorous ablation studies, we meticulously demonstrate the critical role of each module within SQLfuse, validating their combined effectiveness in our system's architecture.
    \item SQLfuse has achieved remarkable success, reaching an execution accuracy of 85.6\% on the prominent Spider Leaderboard, positioning it as the highest-ranked open-source system available to date in the field of Text-to-SQL.
    \item SQLfuse has been successfully integrated into the operational framework of Ant Group, supporting the largest online data analytical processing and transaction processing platforms within the company.
\end{itemize}

\section{Related Work}

In this section, we present an overview of the methodologies applied in Text-to-SQL conversion. The task presents significant challenges that stem primarily from the complexities of accurately interpreting natural language and generating corresponding SQL queries, as extensively documented in recent surveys \cite{qin2022survey,katsogiannis2023survey}. Both database management and natural language processing (NLP) research communities have invested considerable effort in addressing these challenges. Early Text-to-SQL approaches were predominantly based on predefined rules or templates \cite{baik2020duoquest,quamar2022natural,sen2020athena++}. These methods conceptualized the conversion task as a straightforward mapping exercise from natural language to SQL. Other techniques approached the problem from a sequence-to-sequence learning perspective, applying encoder-decoder models to capture the translation process \cite{cai2017encoder,popescu2022addressing,qi2022rasat}. However, recent advancements have seen the emergence of hybrid methods that synergize the strengths of both database and NLP technologies. These include approaches that consider schema relations \cite{hui2022s,li2023graphix,qi2022rasat,wang2019rat,wang2022self,zheng2022hie,liu2023multi} and others that incorporate syntax parsing techniques \cite{guo2019towards,li2023resdsql,scholak2021picard,wang2022proton}. Notably, models based on BERT~\cite{devlin2018bert}, a groundbreaking NLP framework, have been particularly effective in this group, achieving state-of-the-art results in Text-to-SQL conversion \cite{brunner2021valuenet,yin2020tabert}.

Recently, the advent of LLMs like GPT-4 \cite{openai2023gpt} from OpenAI and LLaMA \cite{touvron2023llama} from Meta has marked a significant milestone in NLP and machine learning. Unlike the aforementioned models, LLMs are pre-trained on vast text corpora and can perform diverse language tasks. Their operation is based on generating the next most probable word given the input prompt \cite{zhao2023survey}. In the realm of Text-to-SQL, the efficacy of LLMs largely hinges on the art of prompt engineering—the process of devising precise and effective prompts that guide the model \cite{liu2023comprehensive,nan2023enhancing}. This practice can be bifurcated into two main strategies, dependent on the number of examples supplied to the model: zero-shot and few-shot. In the zero-shot approach, where no examples are provided, the imperative is to craft a question representation that encapsulates all relevant information, including the database schema, to effectively guide the model \cite{chang2023prompt,dong2023c3,liu2023comprehensive,trummer2022codexdb}. Conversely, the few-shot paradigm entails not only the robust representation of the question but also the careful selection and arrangement of a limited number of illustrative examples within the prompt. This approach, known as in-context learning, enables LLMs to recognize and leverage patterns from the provided examples to generate accurate responses, thus equipping them to assimilate new tasks during inference without the need for explicit task-specific training \cite{dong2022survey}. Recent literature highlights the pivotal role of example selection in enhancing the effectiveness of in-context learning \cite{guo2023case,liu2021makes,pourreza2024din}.

On the other hand, while LLMs demonstrate promise in both zero-shot and few-shot scenarios, their proficiency can be further amplified through supervised fine-tuning (SFT)~\cite{sun2023sql}. This method involves additional training on task-specific data, enhancing the model's performance on particular downstream tasks. Recently, SFT has been applied as an alignment training approach, fine-tuning LLMs to mitigate the generation of content that may be offensive, biased, or inaccurate—a critical consideration in deploying these models responsibly.

Despite progress in the field, current LLM-based Text-to-SQL systems have yet to fully leverage the capabilities of LLMs, and they have largely not integrated external tools and collective knowledge to their advantage. As detailed in \autoref{tab:related-works-methods}, these systems often neglect the potential of one-to-many relationships between values in two tables and the association of enumeration values with their natural language counterparts, even though such insights can be gleaned through schema mining to bolster Text-to-SQL performance. Additionally, the majority of existing methods do not employ schema linking, which involves identifying and focusing on the tables and columns pertinent to a user's query—a technique that could significantly aid LLMs in generating more relevant SQL queries. Furthermore, few systems utilize a chain-of-thought (CoT) approach in SQL generation or capture SQL execution messages to rectify errors in the generated queries, and almost none adopt constant value fixes that could automatically correct errors related to constant mismatch in SQL commands. A notable omission in current Text-to-SQL research is the critic module, which has the potential to enhance generation quality by selecting the best SQL statement from a set of candidates. Similarly, the integration of an external knowledge base comprising exemplary Text-to-SQL pairs is not commonly seen in existing works.

SQLfuse distinguishes itself by optimizing the use of LLMs, decomposing the Text-to-SQL process into multiple distinct steps, with LLMs tackling specific tasks at each juncture. In addition to this stepwise approach, our system also skillfully integrates external information and tools such as SQL execution feedback, constant value fixes, and an external knowledge base. This holistic integration ensures that SQLfuse not only capitalizes on the full potential of LLMs but incorporates external knowledge and tools in a seamless manner that enhances the overall system efficacy.

\section{Methodology}

\begin{figure*}
    \centering
    \includegraphics[width=0.9\linewidth]{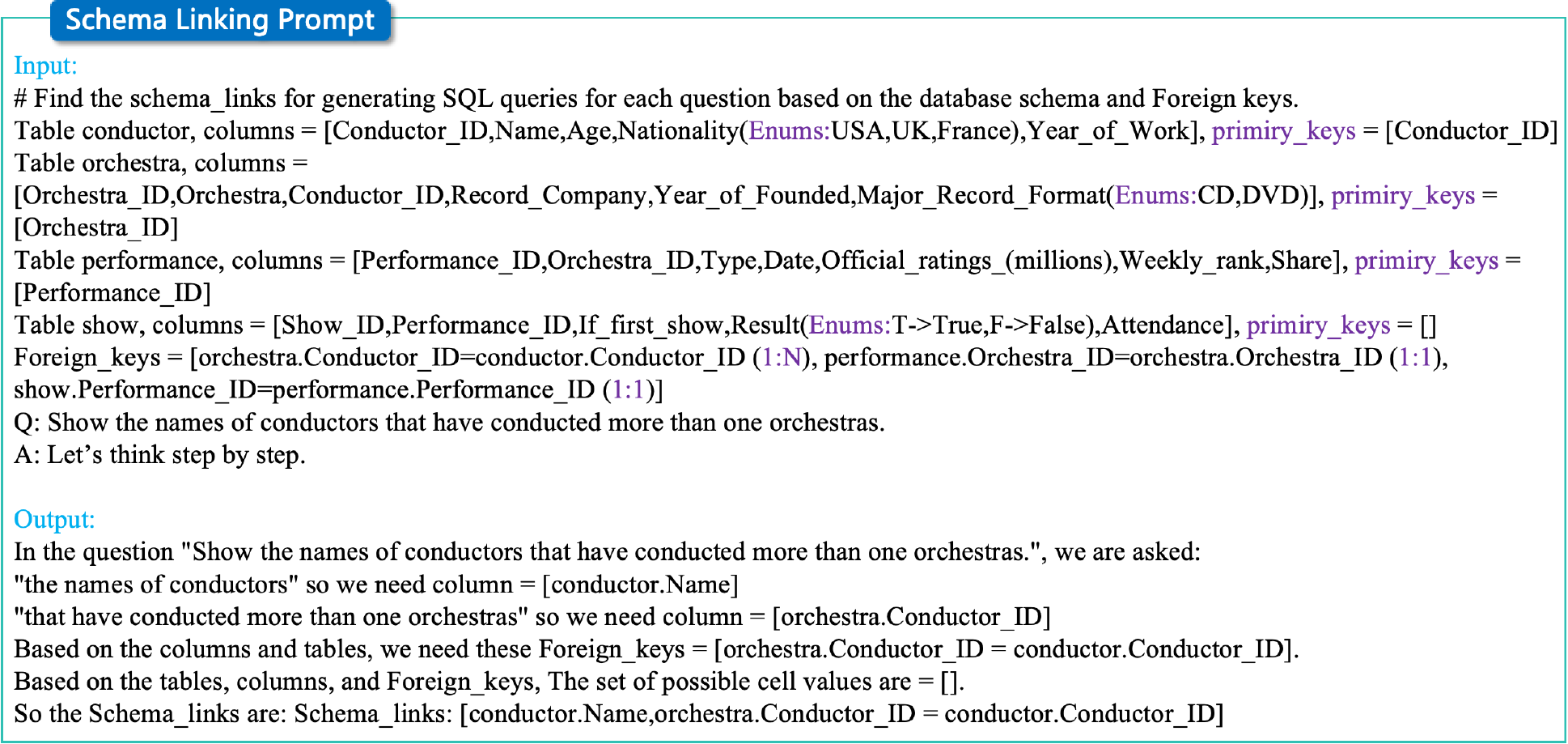}
    %\vspace{-1ex}
    \caption{An example with input and output of schema linking. The contents highlighted in purple color are auxiliary information such as primary keys, enumeration values, one-to-many relations.}
    %\vspace{-1ex}
    \label{fig:schema-linking-example}
\end{figure*}

Our proposed SQLfuse framework, as demonstrated in \autoref{fig:architecture-diagram}, is a modular system designed for Text-to-SQL task. It is mainly composed of the following 4 modules: \protect\circled{1} schema mining, \protect\circled{2} schema linking, \protect\circled{3} SQL generation (SQLgen), and \protect\circled{4} SQL critic. Specifically, schema mining is a service that extracts schema features, e.g., primary keys, foreign keys, enumeration values and one-to-many relations, etc., in addition to a pool of candidate database schemas. Subsequently, schema linking module identifies the exact schema elements namely tables, columns, join relations and condition values referenced in natural language query. Armed with the user's question and the schema features and elements extracted, we then meticulously construct a Chain-of-Thought (CoT) template that serves as a structured prompt for the SQLgen module. This module is responsible for generating a variety of SQL statement candidates, which are subsequently processed through constant value fixing and SQL execution checks for validation. The final stage of the process involves the SQL critic module, which employs few-shot in-context learning to evaluate and select the optimal SQL query that most faithfully represents the user's intent. We will now provide a detailed exploration of each module.

\subsection{Schema Mining Module}
The schema mining module is adeptly crafted to distill essential schema features from databases in a data-driven manner, thereby enriching the context for subsequent modules with vital table information, keys, relationships, and enumeration values. These insights become the cornerstone for both the schema linking and SQLgen modules, allowing them to identify necessary schema components with greater precision and generate SQL statements with heightened accuracy. By leveraging existing knowledge and advanced techniques from database management systems, the schema mining module lays a solid foundation for informed SQL query generation. Now let us delve into the specifics of the extracted schema features.

\subsubsection{Primary Key}
The primary key in a database table is a distinctive identifier, typically a specific column or a set of columns, that ensures each record is unique. This element is fundamental to maintaining the integrity and consistency of data, particularly when conducting aggregation operations. In the realm of Text-to-SQL translation, recognizing and incorporating primary keys is indispensable. They often serve as the axis around which aggregation queries—employing functions like \textit{SUM}, \textit{AVG}, \textit{COUNT}, \textit{MIN}, and \textit{MAX}—are structured, especially when combined with the \textit{GROUP BY} clause. In other words, if the information of primary key are provided in advance, the text-to-sql generation can be more accurate, especially for the aggregation queries.

\subsubsection{Foreign Key}
A foreign key in a database is a field or set of fields that references the primary key of another table, establishing a critical link between them. This link is vital for preserving referential integrity and enabling coherent multi-table queries. Understanding foreign keys is essential for composing SQL queries that require joins across multiple tables.  For instance, to link a <user> table with an <order> table via the field [user ID], the [user ID] in the <user> table acts as the primary key, while in the <order> table it serves as a foreign key. This primary-foreign key relationship illuminates how the tables interrelate, positioning the foreign key as pivotal for the \textit{JOIN} operation. Equipping SQLfuse with foreign key data enriches its capacity to navigate and precisely execute complex multi-table relationships in query translations.

\begin{figure*}
    \centering
    \includegraphics[width=1\linewidth]{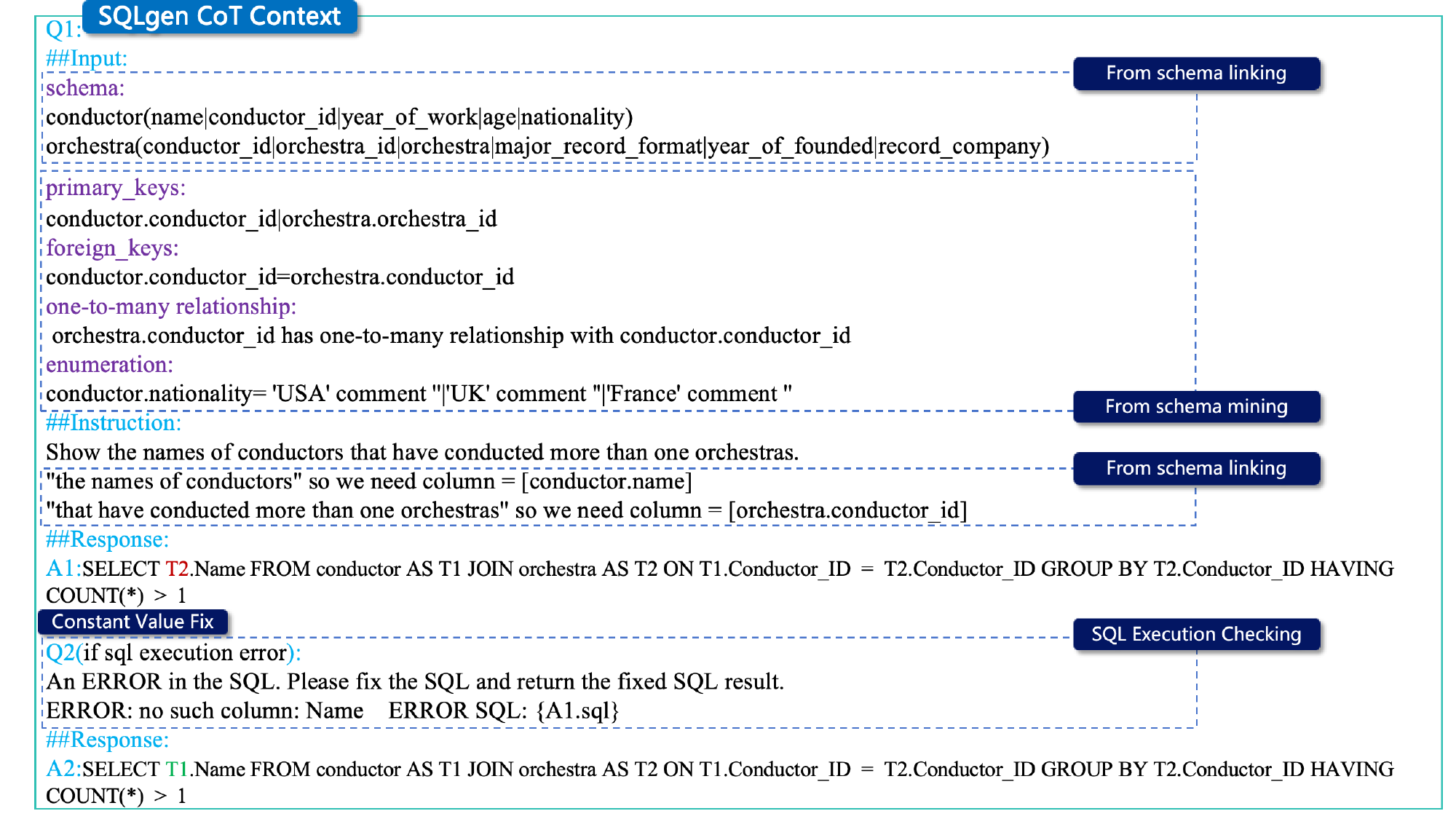}
    \caption{An example illustrates SQLgen prompt with CoT and the generated SQL. It contains a second turn in case when there exist execution errors. }
    \label{fig:sqlgen-illustration}
\end{figure*}

% 在Spider Dev数据集上能够提升 3.4 \% 的准确性
\subsubsection{One-to-Many Relationship}
A one-to-many (1:N) relationship in database design denotes that a single record in one table is associated with multiple records in another. For example, in a <customer> table, each customer (the ``one'' side) may have multiple orders listed in an <orders> table (the ``many'' side). While primary and foreign keys explicitly define inter-table connections, 1:N relationships are often implicit, yet they are essential for understanding data structures and formulating effective SQL queries. In Text-to-SQL translation, accurately identifying these relationships is critical for the proper execution of aggregate queries involving sums, averages, or counts. Recognizing 1:N relationships allows the SQLgen module to make informed decisions on column usage in aggregate operations and the \textit{GROUP BY} clause. Methods to identify these relationships include:
\begin{enumerate}%[leftmargin=1.6em,itemsep= 2pt,topsep = 2pt,partopsep=2pt] 
\item Data value analysis: By scrutinizing actual data, we can detect columns with unique values and those with replicated corresponding entries, hinting at 1:N relationships. 
\item Database statistics: Exploiting statistical data from the database system, like index attributes and column value distribution, can help identify relational patterns. 
\item Machine learning approaches: Machine learning algorithms can classify or cluster table records, revealing hidden data relationships and aiding in the identification of 1:N links. 
\end{enumerate}
Integrating these approaches, the schema mining module becomes adept at discerning the complex web of database relationships, thereby significantly improving the accuracy and logic of the translated SQL queries.

\subsubsection{Enumeration Values}
Enumeration values in a database typically denote a finite set of constants that correspond to a particular column's values. These values might be numeric or alphanumeric codes in the database, but users often refer to them using their full descriptive names or phrases when making queries. For example, a [status] field could be represented by numeric codes such as ``1'' for active and ``2'' for inactive, while users might query these statuses using terms like ``active'' or ``inactive''. Clear semantic mappings between these enumeration values and the terms users employ are crucial for effective schema linking and SQL generation within Text-to-SQL translation. To achieve this, an effective method involves creating mappings that align the database's enumerated codes with their natural language descriptions, using field comments as a guide where available.

%In database design, enumeration values are often used to represent a predefined set of constants corresponding to specific column or field values. In practice, the enumeration values stored in a database might be integers, abbreviations, or codes, while users tend to use full description, more descriptive words or phrases when posing questions. For instance, a field representing status may be stored as numbers in the database (e.g., ``1'' for active, ``2'' for inactive), while users may use natural language terms like ``running'' or ``stopped'' in their queries. In such cases, it is necessary to establish a clear mapping between the database representation and the user’s expression, which helps both schema linking and SQL generation. The most useful method of extracting enumeration values information is building maps between actual enumeration values of database fields and the corresponding natural language comments or descriptions. A simple way is just maintaining a map according to the existing comments of enumeration values even though the comments may be short and abbreviate because LLM is good at understanding semantics. Moreover a better way is to collect or construct synthetic samples with diversity, which enhances the ability of understanding and discriminating of LLMs.

\subsection{Schema Linking Module}
% 总体介绍LLM-based schema linking是怎么实现的
Schema linking plays a crucial role in converting natural language queries into SQL queries, aiming to map input questions to specific database schema elements including tables and columns \cite{li2024codes}. We formulate schema linking as a task of extracting relevant schema items among a set of candidate database schemas. By exploiting the strength of LLMs in natural language understanding, we adopt an LLM-based approach to conduct the task of schema linking. In particular, we construct a decent amount of high-quality labeled data, with which an open-source LLM is further fine-tuned in a supervised manner. Moreover, we manage to enhance the model performance on schema item extraction by supplementing the input with additional context provided by the previous schema mining module. Lastly, the output of schema linking is double-checked and refined with extra schema elements if necessary.

\subsubsection{Schema Items Extraction}
\label{sec:schema-items-extraction}
The training data used for the schema items extraction task stem from a variety of open-source Text-to-SQL datasets. We take these collected and filtered datasets as a basis to pose questions to GPT-4 in a schema-linking prompt style similar to DIN-SQL \cite{pourreza2024din}. In particular, we guide GPT-4 to answer in a Chain-of-Thought (CoT) format by instructing \enquote{Let's think step by step}~\cite{kojima2022large} to incite a thorough elucidation of the extraction rationale. This CoT-guided reasoning is particularly beneficial for the subsequent SQLgen model, enriching its understanding of the semantic ties between user queries and the schema items it extracts, thereby refining the accuracy of SQL statement generation. After acquiring the responses from GPT-4, we examine and verify the correctness of these outcomes through comparison with the schema items, e.g., tables, columns, join relations, and condition values, obtained from the ground-truth labels. By excluding those erroneous question-answer pairs, we end up with a temporary version of high-quality labeled samples. 

Inevitably, there exist columns with abbreviations or sharing identical names across multiple tables. In light of these issues causing ambiguity and impairing the result of schema linking, we refine the quality of our training dataset by adding extra context in the prompt to enable a more precise comprehension. Concretely, we introduce the primary keys, annotations or comments of tables and columns, enumeration values of certain fields, and one-to-many relations between columns.
\autoref{fig:schema-linking-example} illustrates a typical example showing the input and output of schema linking described above.
% 通过采集和过滤开源NL2SQL数据集，以这些数据集为基础，我们通过采用类似dinsql的prompt风格向gpt提问，获得了初始schemalinking训练数据。在得到这些初始数据后，通过工程化手段检查gpt4回答的是否正确，这些工程化的手段包括通过sql解析出投影列、过滤列、排序列、分组列、关联列和gpt4生成的结果进行比对，又进行了进一步的过滤。\\
% 我们评测对比了使用初始数据进行微调的模型效果 和 使用针对性过滤的精选数据进行微调的模型效果，实验结果表明，训练数据的质量对模型的效果影响很大，在训练数据达到一个极高的质量的情况下，模型的效果达到了预期。

Finally, we choose an open-source LLM with strong performance on natural language, namely Llama2-70B, as the schema linking model, and further fine-tune it on the downstream task of schema items extraction based on the constructed dataset. The resulting well-tuned LLM acts as the core of our schema linking module and is capable of extracting the most relevant schema items with respect to user questions.

% In practice, we notice that the quality of schema linking has a significant impact on the performance of subsequent SQL generation. After examining the failed cases of SQL generation, we realize a majority of them can be attributed to either selecting the wrong columns or missing certain columns and relations, due to the existence of non-standard abbreviations or columns sharing identical names across multiple tables. To address such challenges, additional contents are prompted to help the schema linking model understand context. For instance, annotations or comments of tables and columns, enumeration values of certain fields, and one-to-many relation between columns are considered useful auxiliary information and beneficial to improving the performance of SQL generation. 

% 这块就写 起初和sqlgen是做mft，但是对比下来，schemalinking在语言模型上的效果比在代码模型上的效果更好，我们猜测是因为schemelinking任务更考察对自然语言问题的理解能力，后续就和sqlgen的训练任务分开，使用了不同的开源基座模型。

\subsubsection{Schema Items Calibration}

Furthermore, we take extra measures to calibrate the result of schema items extraction. On one hand, we employ a rule-based checking approach to examine if there exists a join relation between two identified tables. The neglected join relation must be restored in the output of schema linking model. On the other hand, we adopt a similarity-based ranking method to recall a limited quantity of potentially relevant candidates from the filtered schema items. Firstly, the annotation and name of a table or column are considered as their sentence representations. By taking advantage of an open-source text embedding model (e.g., text2vec \cite{Text2vec}), we can compare their sentence similarities with the input question. The highly ranked tables and columns above a certain similarity threshold are added to the final results of the schema linking module that serve as the input to the SQL generation module in the next stage.

\begin{figure}[]
%\vspace{-2ex}
\centering
\subfigure[Code Representation Style]{
\centering
\includegraphics[width=1\linewidth]{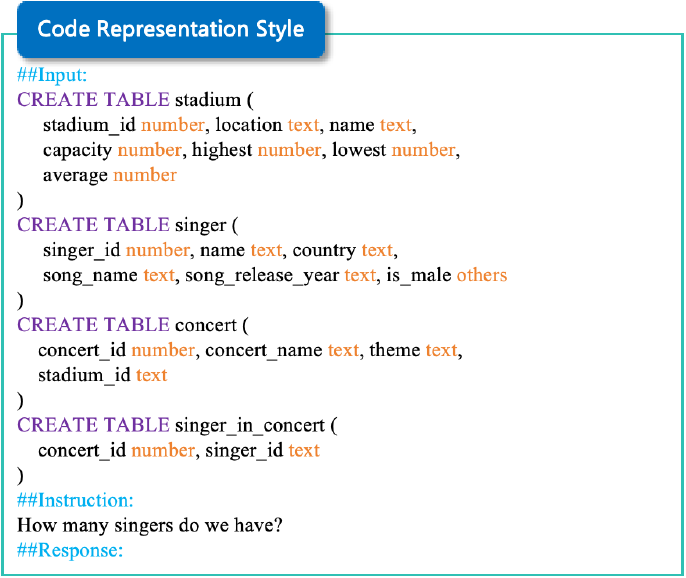}
\label{sfig:code-style}
}\\
\vspace{-1ex}
\subfigure[Natural Language Style]{
\centering
\includegraphics[width=1\linewidth]{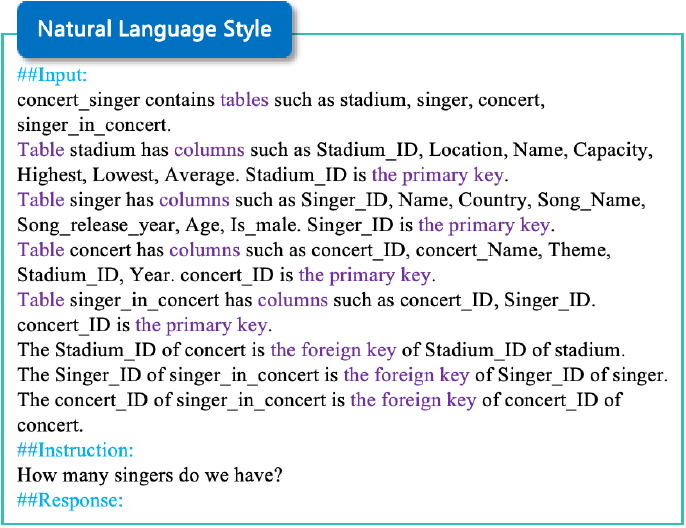}
\label{sfig:nl-style}
}\\
\vspace{-1ex}
\subfigure[SQLfuse Style]{
\centering
\includegraphics[width=1\linewidth]{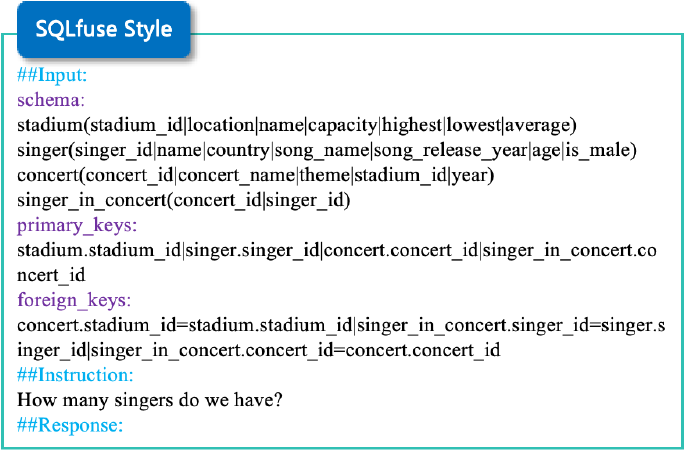}
\label{sfig:sql-gpt}
}
\vspace{-2ex}
\caption{Different prompt styles in SQL generation}
 \label{fig:3_prompt_styles}
\vspace{-3ex}
\end{figure}

\subsection{SQL Generation Module}

% SQLgen（SQL 生成）模块主要分为 4 个方面：

% ● Prompt style：清晰的 Prompt 布局可以帮助模型更好的理解

% ● Chain of Thought：将分析问题的思路告知模型，帮助模型全面思考

% ● Supervised Fine-Tuning：寻找 Text-to-SQL 训练的最优实践

% 最终通过 self-consistent 的方式，输出给 Critic Model 进行选择
As illustrated in \autoref{fig:sqlgen-illustration}, the SQL generation (SQLgen) module serves as the cornerstone of SQLfuse, adeptly synthesizing essential contextual information from schema mining, schema linking, and the original user query. It harnesses a Chain of Thought (CoT) framework, encapsulating these elements in a distinctive SQLfuse style, and then dispatches this amalgamated prompt to a fine-tuned Large Language Model (LLM) to generate SQL predictions. Subsequent to the initial prediction, any type errors in constant values are rectified, and the SQL is executed to identify potential errors for self-correction. This iterative process continues until a valid executable SQL is produced or predefined limits are reached. We will dissect the SQLgen module into two main components: the SQL generation process, and the self-correction process.

\subsubsection{SQL generation}

% 不同 prompt 格式，对于模型理解有不同的效果。甚至，不同模型的最佳 prompt 格式也不一样。经过各个开源模型+各种 prompt 格式实验的对比，我们在开源模型 deepseek 上使用了 Alpaca SFT Prompt，C3 \cite{dong2023c3} 中也探讨过类似的场景，清晰布局的准确率比复杂布局高出 7 \% 。prompt 样例：\autoref{fig:alpaca-style-clear-layout-prompt}
Here, we first introduce the specialized SQLfuse prompt style, which effectively incorporates the context provided by the previous two modules. The format of prompts has been shown to greatly affect a model's understanding, and the optimal format may vary between models. As demonstrated in C3~\cite{dong2023c3}, prompts with a simple and clear layout have been found to outperform those with a more complex structure, increasing accuracy by up to 7\%. Common approaches for Text-to-SQL typically employ either code-centric prompts, which adhere strictly to database schema elements (as depicted in Figure~\ref{sfig:code-style}), or NLP-style prompts, which favor a more versatile, yet less code-oriented format (illustrated in Figure~\ref{sfig:nl-style}).

By conducting comparative experiments with various open-source models and different prompt structures, we have developed a novel prompt style known as the SQLfuse style (visualized in Figure~\ref{sfig:sql-gpt}). This style is adept at representing SQL schema in a compact code-like format while still presenting natural language components, such as the user query, descriptions, and the Chain of Thought process from the schema linking module, in an accessible, conversational manner.

% \begin{figure}
%     \centering
%     \includegraphics[width=1\linewidth]{figures/3-prompts-style.pdf}
%     \caption{Prompt styles with clear layout of schema and data features}
%     \label{fig:3_prompt_styles}
% \end{figure}

%\subsubsection{Chain of Thoughts Context and Supervised Fine-Tuning}

% Chain of Thought是一种模拟人类解决问题时思考过程的方法，它通过提供一系列的中间步骤或推理链，帮助模型理解如何从问题到答案的转换。在Text-to-SQL任务中，CoT可以用来展示Schema Linking模块如何根据问题的内容逐步选择和链接正确的schema元素。

% 通过将CoT信息包含在模型的训练数据中，模型不仅学习到最终的SQL查询语句，还了解到生成这些语句所需的推理过程。例如，对于一个查询“显示所有状态为‘初始化’的配置项”的问题，CoT可能包括以下步骤：

% 1. 确定“配置项”的相关表

% 2. 找到表中表示“状态”的列

% 3. 识别“初始化”状态在数据库中的枚举值表示（如“0”）。

% 4. 构造一个条件语句来过滤出状态为“初始化”的记录。

% 在模型的训练过程中，CoT信息作为额外的上下文来辅助模型学习。这些信息不仅让模型更清楚地了解每个步骤的逻辑，也帮助模型理解复杂查询背后的推理链条。当模型在实际应用中遇到类似的问题时，CoT可以作为内部的指导机制，帮助模型根据问题的内容和数据库的schema进行有效的推理。

% CoT的加入有以下几个优势：

% 1. 提高透明度：CoT使模型解决问题的过程更加透明，使开发者和用户更容易理解模型的决策过程。
% 2. 促进通用性：通过理解解题的逻辑步骤，模型可以更好地处理多变和未见过的查询，提高其泛化能力。
% 3. 增强鲁棒性：当模型生成的SQL语句无法正确执行时，CoT可以帮助定位错误发生在推理的哪个环节，从而提供针对性的修正。

% 总之，CoT信息的加入可以显著提升模型对于复杂Text-to-SQL任务中Schema Linking过程的理解，增强模型的推理能力，从而提高生成SQL查询的准确性和鲁棒性。通过这种方式，可以使模型更加接近人类解决问题的方式，为更高级别的自然语言理解和数据库交互铺平道路。

% 在Spider Dev数据集中，增加CoT信息能够提升 1.1 \% 的准确性

Building on the SQLfuse prompt style, we merge (i) schema and data features (notably primary and foreign keys, one-to-many relationships, and enumeration values) identified by the schema mining module with (ii) the CoT context from the schema linking module, which meticulously matches the correct schema elements to the query content (refer to \autoref{fig:schema-linking-example}). When paired with the corresponding SQL queries from open-source datasets such as Spider~\cite{yu2018spider}, this prompt structure enables us to further fine-tune an open-source LLM to generate high-quality SQL statements within the SQLfuse framework. Specifically, we opt for the leading code-centric LLM, CodeFuse-DeepSeek-33B~\cite{di2023codefuse,liu2023mftcoder}, which, as of March 2024, ranks at the forefront of the Big Code Leaderboard.

\subsubsection{Self-Correction}

To address the inherent output instability of LLMs and improve the performance of SQLfuse, we present a component in SQLgen to rectify common errors through two main methods: constant value fix, and SQL execution checking.

The Constant Value Fix process addresses mismatches between constant values mentioned in natural language queries—like specific numbers, dates, or text strings—and corresponding database columns, which can arise due to ambiguity in language or model misinterpretation. This automatic correction sequence entails:
\begin{enumerate}%[leftmargin=1.6em,itemsep= 2pt,topsep = 2pt,partopsep=2pt] %label=\roman*.,
    \item Constant Value Identification: The system identifies constant values within the query.
    \item Column Matching: It aligns these values with suitable database columns, mindful of data types.
    \item Feedback Verification: Should match attempts falter or provoke SQL errors, a potential mismatch is flagged.
    \item Alternative Matching Exploration: The system then scours for alternative column matches, scrutinizing data formats, value ranges, and metadata.
    \item SQL Modification: The correct column match leads to the immediate adjustment of the SQL by updating the constant value accordingly.
\end{enumerate}
Additionally, database statistics, such as value ranges within columns, can also aid in refining the column matching process. Moreover, heuristic rules tailored to the database's industry-specific terminology and usage patterns can enhance the matching precision, particularly when the data for fine-tuning the above SQL-oriented LLM is insufficient or the scenarios are complex. These strategies not only reduce errors in processing constant values but also improve the overall quality of SQL generation, increase user trust, and boost the model's real-time performance by enabling automatic error correction without manual intervention.

On the other hand, SQL execution checking leverages the the execution error message returned by the database (e.g, the syntax, logical, and database-specific implementation errors) to the erroneous SQL statement and presents it back to the fine-tuned LLM for refinement. This approach does not require additional user interaction for query adjustment, thus maintaining simplicity and general applicability across different databases and use cases. By integrating error-related information from database logs, schema details, or execution context, such as constraints violated or data types involved, the model can pinpoint and understand error causes more accurately, thereby reducing similar future errors. For instance, a type error from attempting numerical operations on a text field informs the model about appropriate data type handling. The integration of the field-type information further refines the model's understanding, leading to more robust error prevention. In addition, this feedback-driven approach, augmented by techniques like Supervised Fine-Tuning (SFT) or reinforcement learning, enables the model to adapt to new challenges and continuously improve its accuracy and user experience over time.

\begin{figure}
    \centering
    \includegraphics[width=0.8\linewidth]{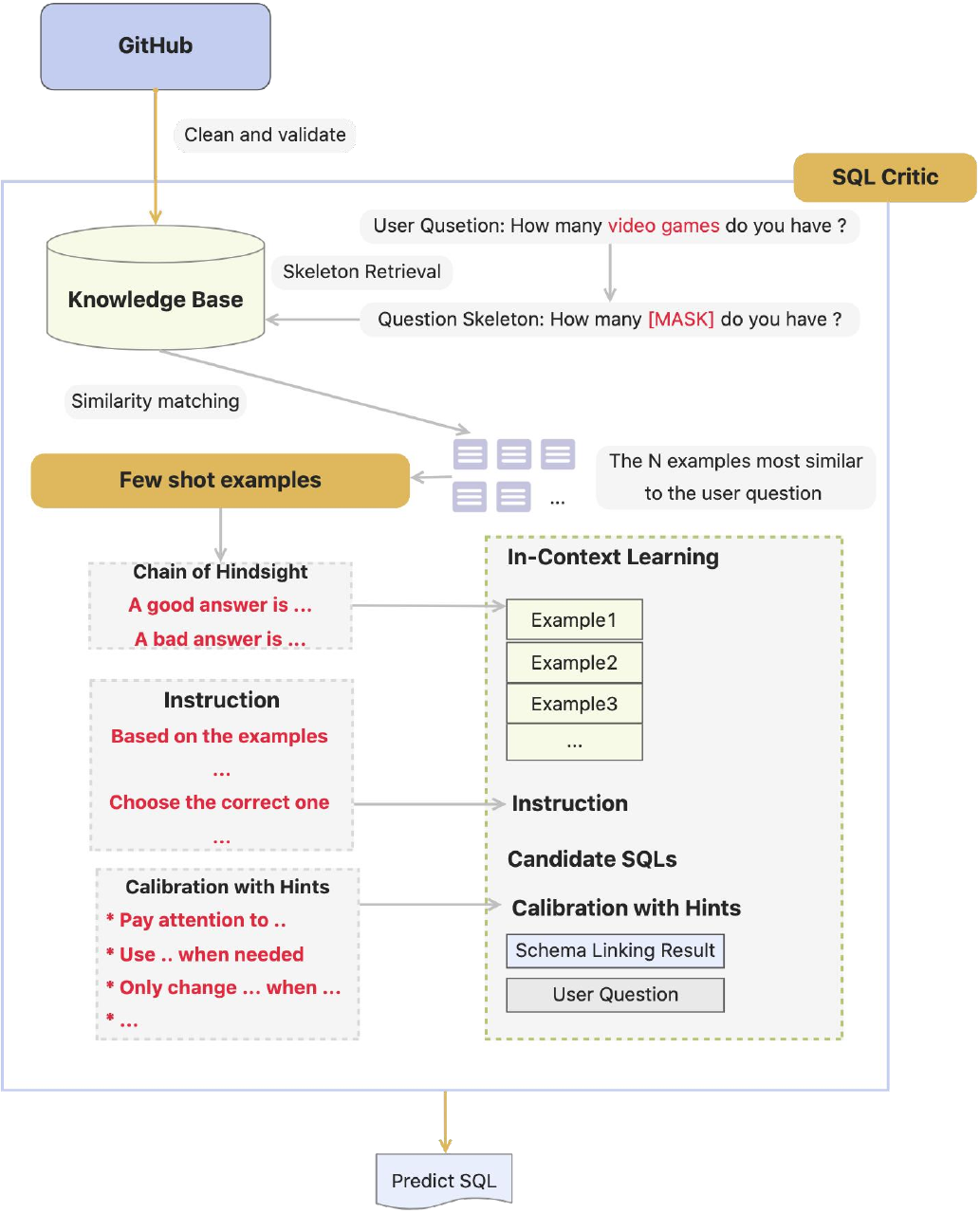}
    \caption{Illustration of SQL critic module}
    \label{fig:critic-module}
\end{figure}

\subsection{SQL Critic Module}

% \begin{figure}
%     \centering
%     \includegraphics[width=1\linewidth]{2.png}
%     \caption{Generate-then-Rank}
%     \label{fig:generate-then-rank}
% \end{figure}

Inspired by the work \cite{pan2023automatically} to further refine the final result of Text-to-SQL generation, we employ a Generate-then-Rank technique, which is a common strategy to correct a LLM at generation time. This technique utilizes a SQL critic model that sifts through possible SQL queries generated by the SQLgen module to identify the most accurate one. The LLM within the SQL critic module is pivotal for assessing SQL quality. Rather than relying on extensive SFT, we adopt a few-shot in-context learning strategy that leverages examples from an external SQL knowledge base, enriched with hindsight feedback. These examples, paired with the user's question, schema linking results, directives, and additional calibration cues, are used to prompt the LLM in the SQL critic module to determine the optimal candidate SQL query, as illustrated in \autoref{fig:critic-module}.

\subsubsection{Few-shot In-Context Learning}
To enable the use of either open-source or closed-source LLMs for the SQL critic model, our system leverages few-shot in-context learning. This method requires only a handful of labeled examples to instruct the LLM, circumventing the need for extensive data collection and fine-tuning. The LLM is thus primed to apply its learned knowledge to novel inputs using this minimal yet effective sample set.

To enrich the variety of cases presented in few-shot learning, we gather and curate an array of intricate SQL statements and schemas from GitHub. This collection undergoes meticulous cleaning and validation, including manual annotation and GPT-4 evaluation, resulting in a robust external knowledge base. This repository enables the retrieval of historically similar samples to the input question through a method that prioritizes similarity in retrieval, as depicted in \autoref{fig:critic-module}. We further refine the retrieval process by masking specific keywords in the question, retaining only its \enquote{question skeleton}, which enhances the ability to abstract away domain-specific details and fosters more generalized sentence comparison.

The few-shot examples are also enriched with a Chain of Hindsight (CoH) \cite{liu2023chain} framework, which incorporates feedback prompts like \enquote{A good answer is} and \enquote{A bad answer is} (see \autoref{fig:critic-module}). This technique informs the model using retrospective insights, enhancing its capacity to generate high-quality SQL outputs.

\begin{figure}
    \centering
    \includegraphics[width=0.6\linewidth]{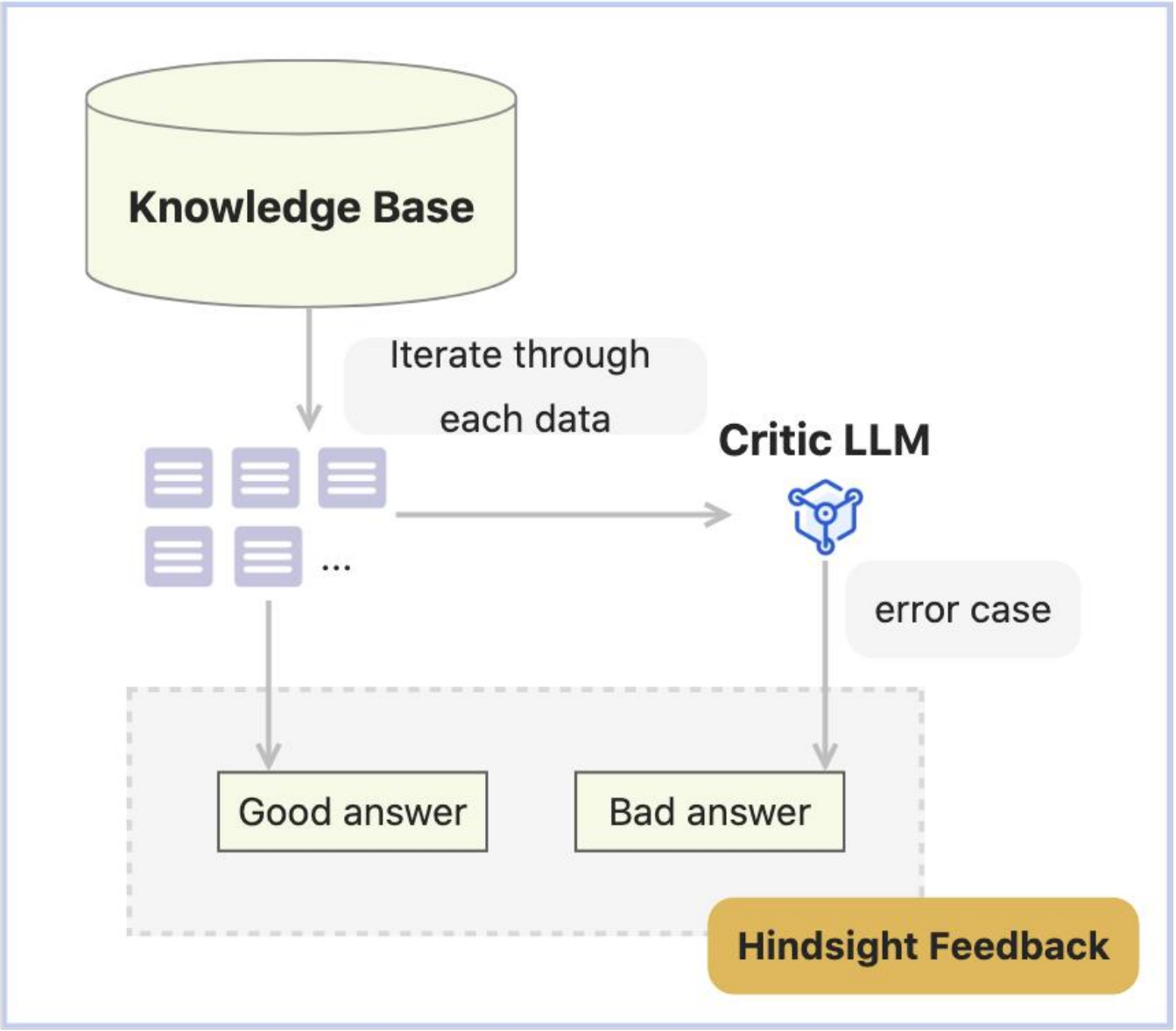}
    \caption{Generating hindsight feedback}
    \label{fig:coh-example}
\end{figure}

One challenge with using historical SQL queries from the knowledge base is that they typically feature a single correct answer without the nuanced feedback often provided by human annotators. To facilate the CoH feedback with negative answers, the LLM within the SQL critic module is engaged to generate suboptimal answers, adding context to the knowledge base. By labeling these inferior responses as "bad answers" in juxtaposition with the "good answers" represented by the original labels, we equip the model to discern quality, as showcased in \autoref{fig:coh-example}.

Finally, the few-shot examples are also complemented by instructions that explicitly direct the SQL critic model to identify the correct SQL query from the array of options provided by the SQLgen module. This intentional placement of a clear command is key to guiding the model's selection process effectively.

\subsubsection{Instruction and Calibration Hints}

To further refine the decision-making capabilities of the critic model, the input prompt is supplemented with calibration hints that enumerate common errors typically made by the SQL critic module. These hints serve to preemptively address potential missteps, enhancing the model's ability to generate a superior SQL query. The calibration hints, combined with the user's question and schema linking results, form a comprehensive prompt designed to steer the model toward producing the most accurate SQL output, as illustrated in \autoref{fig:critic-module}.

\section{Experiments}

\subsection{Dataset}

% Spider 数据集 是一个大规模的跨领域 Text-to-SQL 基准数据集，Spider 数据集包含了 8,659 个训练样本，包括问题和 SQL 查询对，以及 1034 个开发样本，涵盖了 138 个不同领域的 200 个复杂数据库。Spider 的提出使人们将 Text-to-SQL 任务的注意力转向“在没有见过的数据库结构上模型的泛化” ( generation to unseen database schema ) 问题。这个泛化问题的挑战主要有三方面：\\
% ● Schema linking 的难度增加：如何在未见过的 Schema 背景中，分析找到对应 question 应该使用的表和列；\\
% ● 在 SQL 生成阶段，如何组织已有的信息，甚至结合数据库中的数据，来帮助模型更好的理解；\\
% ● 如何检索组织外部知识库去弥补模型泛化性的不足。\\

We select the Spider benchmark~\cite{yu2018spider} in all of our experiments for the reason that Spider is the most difficult and influential Text-to-SQL benchmark among all existing benchmarks of authority. The Spider dataset encompasses a substantial collection of cross-domain queries. It includes a training set with 8,659 samples (TRAIN Set) comprising question-and-SQL-query pairs, alongside 1,034 development samples (DEV Set) that span 200 intricate databases across 138 distinct domains. Additionally, for competitive assessment, an undisclosed test dataset (TEST Set) is offered to facilitate leaderboard rankings. The advent of Spider has steered research efforts toward addressing the critical issue of model generalization across novel and unseen database structures. In light of Spider's authoritative stance within the field, we have conducted our experimental evaluations on its DEV and TEST sets, with a focus on SQL accuracy—measuring both the execution viability and the correctness of the resultant data output.

\subsection{Experimental Setups}
We introduce the LLMs used in schema linking, SQL generation and SQL critic module.

\begin{itemize}%[leftmargin=1.5em,itemsep= 2pt,topsep = 2pt,partopsep=2pt]
\item{Schema Linking Model}:
An open-source LLM Llama2-70B is selected as the base model of our schema linking module. 

\item{SQL Generation Model}: We choose CodeFuse-DeepSeek-33B, which ranks first on Big Code Models Leaderboard until March 2024, as the base model for SQL generation.

\item{SQL Critic Model}:
An open-source LLM Llama2-70B is mainly used as the SQL critic model. To notice, we also experiment with GPT-4 as the LLM in SQL critic module.  
\end{itemize}

\subsubsection{Fine-tuning Schema Linking Model}
The schema linking model is further fine-tuned with the constructed dataset described in section \ref{sec:schema-items-extraction}. The training dataset consists of $5k$ labeled samples. We adopt a parameter-efficient fine-tuning approach of QLoRA with a rank of $64$. We use a cosine scheduler and AdamW optimizer with a learning rate of $5e^{-5}$ and weight decay of $0.1$. The model is trained with a batch size of $2$ using a single A100 GPU for $8$ epochs. 

Afterwards, we also build a test set to evaluate the performance of the schema linking model based on the DEV set of the Spider benchmark. Specifically, we use recall of table and column as the evaluation metrics. The resulting schema linking model is able to achieve a $99.8\%$ and $97.4\%$ recall of table and column respectively. 
%The best checkpoint produced during the training process is kept as the final model for schema linking task. 

\subsubsection{Fine-tuning SQL Generation Model}
We choose CodeFuse-DeepSeek-33B, an open-source LLM with exceptional capabilities in handling code-related tasks, as our foundation model for the SQLgen module. It is fine-tuned via QLoRA with the TRAIN set from the spider benchmark. The rank of QLoRA is set to be $96$. We use a cosine scheduler and AdamW optimizer with a learning rate of $2e^{-4}$ and weight decay of $0.1$. The model is trained with a global batch size of $16$ using four A100 GPUs for $10$ epochs.
%To be consistent, comparable and competitive, all results are evaluated on the same foundation model CodeFuse-Deepseek-33B, which ranks first in win rate on the Big Code Models Leaderboard as of Feburary 2024, demonstrating exceptional capabilities in handling code-related tasks. All SQLgen-LLMs are fine-tuned via LoRA or QLoRA. The training dataset consists of $9k$ high-quality labeled samples. We adopt a parameter efficient fine-tuning approach of QLoRA with a rank of $96$. We use a cosine scheduler and AdamW optimizer with a learning rate of $2e^{-4}$ and weight decay of $0.1$. The model is trained with a batch size of $4$ using four A100 GPUs for $10$ epochs. 

\begin{figure}[t]
    \centering
    \includegraphics[width=1\linewidth]{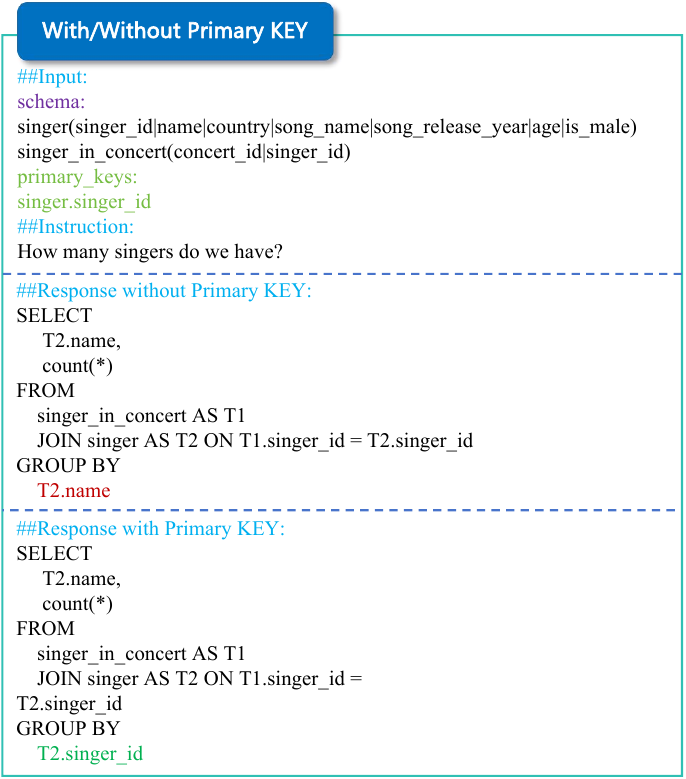}
    \caption{Prompts and responses w/wo primary-key}
    \label{fig:primary-key}
\end{figure}

\begin{figure}[t]
    \centering
    \includegraphics[width=1\linewidth]{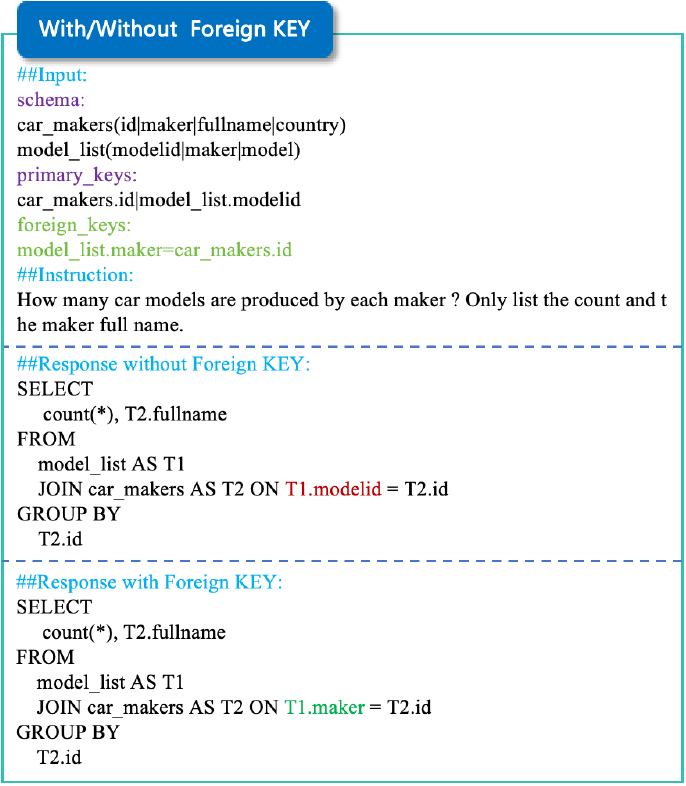}
    \caption{Prompts and responses w/wo foreign-key}
    \label{fig:foreign-key}
\end{figure}

%  SQLfuse整体的榜单性能和成绩
\subsection{SQLfuse Evaluation}

\subsubsection{Spider Leaderboard}
We evaluate the performance of our proposed SQLfuse on the Spider benchmark. Our focus is on assessing the execution accuracy of SQL queries with values on the TEST dataset. According to \autoref{tab:leaderboard}, SQLfuse achieves an impressive $85.6\%$ accuracy, ranking just below MiniSeek and DAIL-SQL \cite{gao2023texttosql}. Notably, MiniSeek and DAIL-SQL, along with other top contenders, primarily utilize closed-source LLMs like GPT-3.5/4 for their operations. Despite this, SQLfuse demonstrates robust competitiveness against these closed-source models and markedly outperforms alternatives that rely on open-source LLMs.

%Specifically, we measure the accuracy of SQL execution with values on the TEST set. According to \autoref{tab:leaderboard}, SQLfuse achieves a $85.6\%$ accuracy following MiniSeek and DAIL-SQL \cite{gao2023texttosql}. Besides, it is noticeable that a majority of the top-ten solutions rely on closed-source LLM, i.e., GPT-3.5/4. In contrast, our proposed SQLfuse is still adequately competitive and outperforms other open-source LLM-based approach by a large margin. 

\subsubsection{Real-world Application}
It is also worth of mentioning that SQLfuse has been tested and deployed in real-world scenarios. Notably, it is integrated into the daily operational framework of Ant Group, supporting seven business contexts including the online analytical processing (OLAP) and transaction processing (OLTP) platforms within the company.

\begin{table*}
    \small
    \centering
    \begin{tabular}{|c|c|c|c|c|}
        \hline
       Rank. & Solution &  TEST Acc &  Type &  LLM\\\hline
       1. & MiniSeek &   91.2 & Unknown & Unknown \\\hline
       2. & DAIL-SQL + GPT-4 + Self-Consistency\cite{gao2023texttosql} & 86.6  &  Prompting & Closed source	 \\\hline 
       3. & DAIL-SQL + GPT-4 \cite{gao2023texttosql} & 86.2  &  Prompting & Closed source	 \\\hline 
       4. & \textbf{SQLfuse} & \textbf{85.6}  &  \textbf{Fine-tuning} & \textbf{Open source} \\\hline
       5. & DPG-SQL + GPT-4 + Self-Correction & 85.6  &  Prompting  & Closed source \\\hline
       6. & DIN-SQL + GPT-4\cite{pourreza2024din} & 85.3  &  Prompting & Closed source \\\hline
       7. & Hindsight Chain of Thought with GPT-4 & 83.9  &  Prompting & Closed source \\\hline
       8. & C3 + ChatGPT + Zero-Shot\cite{dong2023c3} & 82.3  &  Prompting & Closed source\\\hline
       9. & Hindsight Chain of Thought with GPT-4 and Instructions & 80.8  &  Prompting & Closed source\\\hline
       10. & RESDSQL-3B + NatSQL\cite{li2023resdsql} & 79.9  &  Fine-tuning  & Open source\\\hline
    \end{tabular}
    \caption{Rankings on Spider Leaderboard-Execution with Values by Dec. 2023}
    \label{tab:leaderboard}
\end{table*}

\subsection{Ablation Study}
% SQLfuse contains many modules and each module have one or more components/methods. We are going to provide the most important results of ablation study.
SQLfuse consists of four modules and each module employs various techniques to aid a better SQL generation. We conduct different ablation studies to evaluate our approach with and without each of the following modules or methods.

\begin{figure}[t]
    \centering
    \includegraphics[width=1\linewidth]{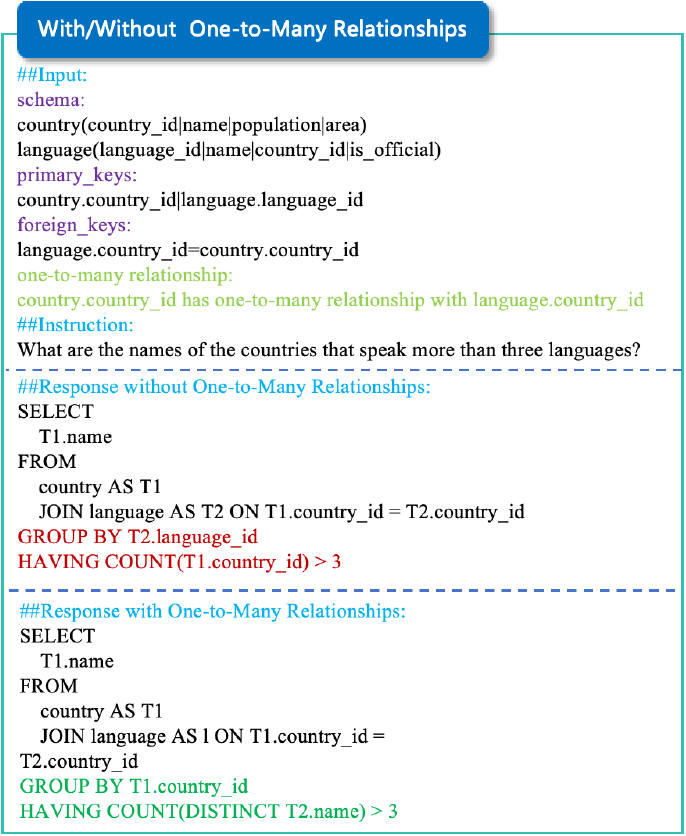}
    \caption{Prompts and responses w/wo one-to-many relations}
    \label{fig:one-to-many}
\end{figure}

\begin{figure}[t]
    \centering
    \includegraphics[width=1\linewidth]{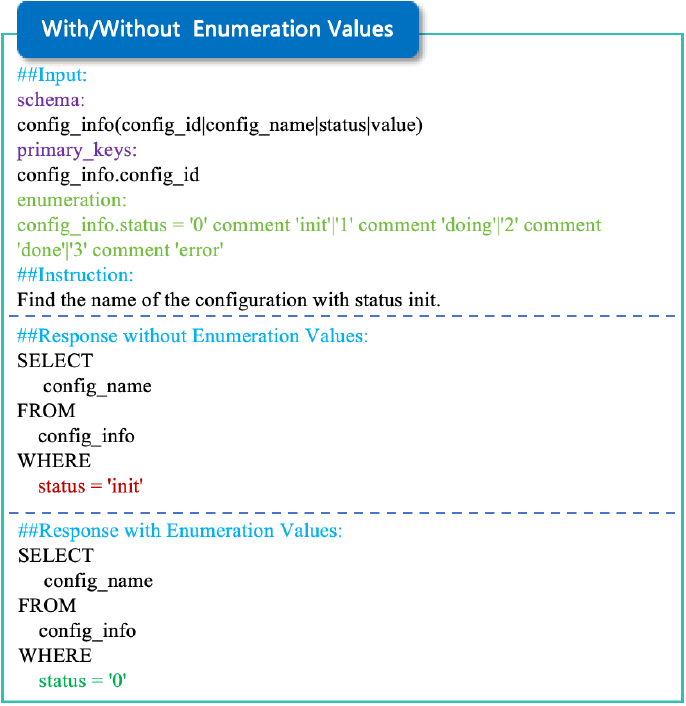}
    \caption{Prompts and responses w/wo enumeration values}
    \label{fig:enum}
\end{figure}

\subsubsection{Schema Mining}

Regarding schema mining, we extract and leverage critical schema attributes to enrich the context provided to our model. We have evaluated the incremental benefits these schema features offer.

\autoref{tab:ablation-study} sheds light on the tangible accuracy losses in the TEST dataset when omitting various schema-related enhancements. To clarify:
\begin{itemize} %[leftmargin=1.5em,itemsep= 2pt,topsep = 2pt,partopsep=2pt]
\item \textbf{Impact of Primary Key:} Excluding primary key cues results in a $1.8\%$ decline in accuracy, underscoring the value of primary key identification for the model’s proficiency. For instance, as depicted in \autoref{fig:primary-key}, confusion may arise in distinguishing between the [singer ID] in a <singer> table and a <concert> table. Recognizing the [singer ID] as the primary key is crucial for generating accurate \textit{GROUP BY} fields in queries such as \enquote{the number of concerts per singer}, ensuring uniqueness and preventing errors.

\item \textbf{Impact of Foreign Key:} The accuracy takes a $3.5\%$ hit without foreign key insights, emphasizing the foreign key's role in multi-table queries. As illustrated in \autoref{fig:foreign-key}, the model, aided by foreign key information, can correctly identify [model id] for joins between <model\_list'> and <car\_maker> tables. Understanding table relationships via foreign keys is pivotal for precise multi-table join operations.

\item \textbf{Impact of one-to-many relationships:} A $1.8\%$ accuracy loss is observed when one-to-many relations are neglected. An example in \autoref{fig:one-to-many} demonstrates the importance of recognizing the one-to-many link between [country\_id] in <country> and <language> tables for queries like \enquote{Show the names of conductors who have conducted more than one orchestra}. Accurate \textit{GROUP BY} choices, such as [country\_id], hinge on this understanding to prevent improper aggregate counts.

\item \textbf{Impact of Enumeration values:} Enumeration value information yields a $0.9\%$ accuracy enhancement. Without it, as \autoref{fig:enum} exemplifies, incorrect 'status' values like 'init' may surface in the \textit{WHERE} clause. Conversely, access to enumeration mappings ensures the correct '0' status is applied, emphasizing the necessity for accurate enumeration contextualization.

\end{itemize}
Each schema feature individually enhances SQLfuse's performance, showcasing their distinct contributions. Collectively, these schema attributes significantly contribute to the holistic efficacy of SQLfuse, as reflected in its overall performance.

\begin{table}
    \small
    \centering
    \begin{tabular}{|c|c|c|c|}
        \hline
       Scenario & DEV Acc & TEST Acc & Diff in Test Acc\\\hline
        SQLfuse & 86.6  &  85.6  & 0.0\\\hline
        wo Primary KEY & 83.6  &  83.8 & -1.8\\\hline 
        wo Foreign KEY & 81.1  &  82.3 & -3.3\\\hline
        wo One-to-Many Relation & 84.6  &  83.8& -1.8\\\hline
        wo Enum Values & 85.5  &  84.7& -0.9\\\hline
        wo Schema Linking & 84.4  &  83.1 & -2.5\\\hline
        wo CoT & 86.3  &  85.5 & -0.1\\\hline
        Code Representation Style & 86.5  &  85.0 & -0.6\\\hline
        Natual Description Style & 85.7  &  85.1 & -0.5\\\hline
        wo Constant Value Fix & 86.5  &  85.2& -0.4\\\hline
        wo SQL Execution Checking & 86.3  &  85.6& -0.0\\\hline
        wo Critic Module & 84.6  &  83.8 & -1.8\\\hline
        Critic Module (GPT-4) & 87.7  &  85.9 & +0.3\\\hline
    \end{tabular}
    \caption{Performance on SQL generation w/wo different modules and techniques.}
    \label{tab:ablation-study}
\end{table}

\subsubsection{Schema Linking}

To understand the significance of schema linking in SQL generation, we evaluate SQLfuse's performance when this module is removed. Consequently, the SQLgen module receives a compilation of potential database schemas and their associated schema features straight from the schema mining module, without the benefit of schema linking. According to the data presented in \autoref{tab:ablation-study}, there is a notable accuracy drop—over $2\%$—on both the DEV and TEST datasets when schema linking is absent. This decline highlights the crucial role of schema linking in identifying pertinent schema components, which is instrumental in producing more accurate SQL queries.

Following this, we investigate the performance implications for SQLfuse without the use of Chain of Thought (CoT) reasoning. In this scenario, the schema linking module provides only fundamental details without the explanatory steps that CoT entails. As per the findings in \autoref{tab:ablation-study}, the marginal gain from CoT is $0.1\%$, signifying that the inclusion of CoT reasoning by itself enhances the model’s output. 

% \begin{table}
%     \centering
%     \begin{tabular}{|c|c|c|}
%         \hline
%        Schema Linking Module & DEV Acc & TEST Acc\\\hline
%          With Schema Linking Module & 84.6  &  83.8\\\hline
%          Without Schema Linking Module & 82.1  &  81.1\\\hline
%     \end{tabular}
%     \caption{Performance w/wo Schema Linking}
%     \label{tab:schema-linking}
% \end{table}

\subsubsection{SQLgen Module}
Our analysis within the SQLgen module continues with a focus on the effects different techniques have on SQL generation proficiency.

Initially, we evaluate the effectiveness of diverse prompt styles, which form the basis of our training data. The base LLM is fine-tuned using these styles to allow for comparative analysis. \autoref{tab:ablation-study} indicates that prompt styles yield varied outcomes, with the SQLfuse style surpassing both the code representation style (which results in a drop of $0.6\%$ in Test set accuracy) and the natural language style (which shows a drop of $0.5\%$ in Test set accuracy), as measured on the Spider benchmark. This superiority is likely due to SQLfuse's advantageous blend of SQL schema representation with natural language elements.

Moving on, we assess the Constant Value Fix and SQL Execution Checking modules. As detailed in \autoref{tab:ablation-study}, both modules contribute to an absolute increase of $0.4\%$ in final accuracy, showcasing their individual significance in enhancing SQL generation.

Lastly, we delve into the fine-tuning methodology of the SQLgen module. Our fine-tuning relies on standard, parameter-efficient techniques, specifically LoRA or QLoRA. Within these methods, the LoRA rank is important as it manages the proportion of trainable parameters and influences the effectiveness of fine-tuning. Distinct LoRA ranks correspond to varying levels of performance. Comprehensive evaluations reveal that a LoRA rank of 96 is the sweet spot for our SQLgen, outperforming both ranks 32 and 228. While increasing the rank initially benefits performance, there is a threshold beyond which further elevation does not translate to better results and might even be detrimental. This is anticipated, given our training dataset encompasses fewer than 10,000 samples, meaning a rank of 96 suffices for our purposes. Note that these accuracy figures pertain to the SQLgen module specifically, rather than the SQLfuse system as a whole.

\begin{table}
    \centering
    \begin{tabular}{|c|c|c|}
        \hline
        LoRA Rank & DEV Acc & TEST Acc\\\hline
         288 & 83.1  &  83.5\\\hline
         \textbf{96} & \textbf{84.6}  & \textbf{83.8} \\\hline
         32 & 82.8 &  82.9\\\hline
    \end{tabular}
     \caption{Performance of SQLgen with Different LoRA Ranks.}
    \label{tab:lora_ranks}
\end{table}

\subsubsection{SQL Critic Module}
Our attention turns to the SQL critic module, evaluating its influence by removing it from the SQLfuse framework and analyzing the performance based solely on the output from the SQLgen module. \autoref{tab:ablation-study} reveals that removing the critic module results in a reduction of approximately $2\%$ in both DEV and TEST set accuracy. This data underscores the critic module's role in refining SQL generation, highlighting its effectiveness in enhancing the quality of the generated queries.

Further, we undertake a comparative study by swapping out Llama2-70B for GPT-4 in the critic's role. This substitution leads to a marginal increase in TEST set accuracy, nudging it from $85.6\%$ to $85.9\%$. Such an outcome indicates that while GPT-4 offers slight improvements, the SQL critic module's structure and functionality are significant contributors to SQLfuse's overall performance, irrespective of the underlying language model.

\section{Conclusion}
In this paper, we present SQLfuse—a cutting-edge contribution to the Text-to-SQL field that represents a substantial leap forward. Our systematic approach, which integrates schema mining, schema linking, SQL generation (SQLgen), and a critic module, is meticulously designed to improve the accessibility and precision of generating SQL queries from natural language. SQLfuse harnesses the expansive capabilities of LLMs while also integrating ancillary knowledge and tools, resulting in a synergy that significantly enhances system performance. The system's landmark achievement of an $85.6\%$ execution accuracy on the renowned Spider Leaderboard serves as a testament to the efficacy of SQLfuse. Ranking as the top open-source LLM-based Text-to-SQL system, SQLfuse extends the frontiers of scholarly research and meets the practical demands of the industry with powerful, user-friendly data querying capabilities. SQLfuse's theoretical strengths are brought to life in real-world settings, as demonstrated by its successful deployment within the dynamic operational frameworks of Ant Group. This practical application confirms SQLfuse's role as an indispensable asset, adept at navigating and managing complex data queries, and solidifying its position at the intersection of academic innovation and industrial utility.

\section{DISCUSSION}

In this section, we explore strategies to further elevate the performance of LLM-based Text-to-SQL translation. Our discourse encompasses techniques currently under experimentation as well as those slated for future investigation. 

\subsection{Training Data Expansion}
In the Text-to-SQL domain, training data expansion is an effective strategy for enhancing model performance, particularly when training data is scarce. We have explored various data augmentation strategies for your reference and discussion. 

A creative method for data expansion is using error set expansion, which expand more data via summarize common errors of model during evaluation on validation dataset. This approach involves analyzing the mistakes of model output, identifying similar issues within the database, and generating new training samples grounded in these erroneous instances. This targeted improvement helps the model to avoid similar mistakes in the future by reinforcing its knowledge on specific points.

The exploration of this data augmentation method entails the following steps:
\begin{itemize}%[leftmargin=1.5em,itemsep= 2pt,topsep = 2pt,partopsep=2pt]
\item Error Analysis Initially: An assessment of the model's errors during evaluation is conducted to identify weak points, such as JOIN operations, subqueries, or aggregate functions.
\item Question Skeleton Generation: To discover structurally similar questions to the error instances, question skeletons are created by abstracting the main structure of the original queries and masking specific values and keywords.
\item SQL Skeleton Generation: Similar to question skeletons, SQL skeletons maintain the structure of the original SQL queries but with specific values and database entity names removed, preserving their structured pattern.
\item Knowledge Base Matching: The generated question and SQL skeletons are used to find structurally similar questions and corresponding SQL statements in a knowledge base, which can be done using text similarity algorithms or models specifically designed for text matching.
\item GPT-based Knowledge Expansion: Once similar questions to the error instances are identified, pre-trained LLMs like GPT are employed to generate new pairs of questions and SQL statements. This method allows for the expansion of the model's knowledge, leveraging GPT's generative capabilities to create new training samples.
\item Data Cleaning and Validation: The newly generated question and SQL pairs must undergo cleaning and validation to ensure their quality is sufficient for effective model training. This process often involves syntax checks, logical consistency validation, and ensuring SQL statements conform to the database schema.
\end{itemize}
Through this error-knowledge-based data augmentation, the model can improve by learning from its weaknesses, offering a targeted solution to rectify deficiencies in handling certain types of problems. This method not only strengthens the model's understanding of existing question types but also equips it to handle new and unseen problem types, thereby enhancing its robustness and generalization capabilities.

Another method of data expansion is complex SQL expansion.
In Text-to-SQL tasks, particularly for applications involving complex queries, it becomes crucial to enhance the model's understanding of intricate SQL structures. This enhancement can be achieved by incorporating a higher number of complex SQL samples into the training data. The following is a general process for augmenting with complex SQL:
\begin{itemize}%[leftmargin=1.5em,itemsep= 2pt,topsep = 2pt,partopsep=2pt]
\item Data Collection: A vast array of complex SQL queries and corresponding schema, often including multiple JOIN operations, nested queries, intricate WHERE clauses, GROUP BY and aggregate functions, are collected from open-source platforms like GitHub.
\item SQL Parsing: The collected SQL statements are parsed to break them down into components, such as SELECT fields, FROM clauses, JOIN conditions, and WHERE conditions, aiding in a better structural understanding of the SQL statements.
\item Schema Matching: Parsed SQL structures are matched with corresponding schema, ensuring the SQL queries reflect the schema's structure by correctly using table and field names.
\item Question Generation: The parsed SQL structures and matched schema are provided to pre-trained LLMs like GPT to generate corresponding natural language questions. This process can be guided by providing templates or prompts to direct GPT in generating questions semantically consistent with the SQL queries.
\item Secondary Check: The questions generated by GPT, along with the original SQL and schema, are put back into GPT for verification, ensuring the generated questions correlate with the SQL queries and accurately reflect the information in the schema.
\item Data Cleaning and Validation: As auto-generated questions and SQL may contain errors or inconsistencies, manual or automated cleaning and validation are required. This may include checking the naturalness of the language in the questions, the accuracy of SQL syntax, logical consistency, and correct usage of schema features.
\end{itemize}
Such a process for complex SQL expansion can effectively create a rich training dataset, reflecting the complexity of real-world SQL queries with corresponding natural language questions. This data is vital for improving the model's ability to handle complex queries, helping it better understand and generate SQL statements that involve sophisticated database operations. Additionally, this method promotes the model's adaptability and generalization in addressing real-world problems, thus enhancing user experience.

\subsection{Preference Learning}
Preference learning such as Reinforcement Learning from Human Feedback(RLHF) and Direct Preference Optimzation(DPO) may enhance the model's ability to generate complex SQL statements. This approach may lead generated SQLs generated more in line with human writing habits, beneficial for both the model's practicality and user satisfaction. We are going to discover how such preference learning influence Text-to-SQL results in the future.

% \begin{acks}
%  This work was supported by the [...] Research Fund of [...] (Number [...]). Additional funding was provided by [...] and [...]. We also thank [...] for contributing [...].
% \end{acks}

%\clearpage

\bibliographystyle{ACM-Reference-Format}
\bibliography{sample}

% \appendix

% \section{Calibration Hints}
% \label{appendix:calibration-hints}
% We provide a list of calibration hints that are summarized and refined by analyzing the cause of failure or error cases of SQL generation. In practice, one or more hints can be added to the input of LLMs to aid in a better SQL query generation accordingly.
% \begin{itemize}
%     \item Use the database values that are explicitly mentioned in the question
%     \item Pay attention to the columns that are used for the JOIN by using the Foreign keys.
%     \item Use DESC and DISTINCT when needed
%     \item Pay attention to the columns that are used for the GROUP BY clause.
%     \item Pay attention to the columns that are used for the SELECT clause.
%     \item Only change the GROUP BY clause when necessary.
%     \item Please choose tables and columns according to the question.
%     \item Must pay attention to the one-to-many relationship.
% \end{itemize}

\end{document}